# Multi-Labeled Classification of Demographic Attributes of Patients: a case study of diabetics patients

Naveen Kumar Parachur Cotha[1] and Marina Sokolova[1,2]

[1]University of Ottawa,
[2]Institute for Big Data Analytics

{npara051,sokolova}@uottawa.ca

## Abstract:

Automated learning of patients' demographics can be seen as multilabel problem where a patient model is based on different race and gender groups. The resulting model can be further integrated into Privacy-Preserving Data Mining, where they can be used to assess risk of identification of different patient groups.

Our project considers relations between diabetes and demographics of patients as a multi-labelled problem. Most research in this area has been done as binary classification, where the target class is finding if a person has diabetes or not. But very few, and maybe no work has been done in multi-labeled analysis of the demographics of patients who are likely to be diagnosed with diabetes. To identify such groups, we applied ensembles of several multilabel learning algorithms. The best performing multi label ensembles include BR/Hoeffding Tree, CC/Hoeffding Tree, BCC/Hoeffding Tree, BR/JRIP, CC/JRIP, BCC/ JRIP respectively. In the empirical part of this study, we used on the UCI Diabetics dataset of over 100,000 records, collected from 130 US hospitals. The dataset consisted of attributes that included personal demographics, diagnoses code, lab results, etc. Experiments conducted on datasets of 1000, 10000, 20000 samples, show that BR/JRip model achieves a high overall accuracy of 0.533 (1000 samples), 0.702 (10000 samples), 0.569 (20000 samples), improving over the baseline model ZeroR with accuracy of 0.526, 0.586, .562 respectively. Loss functions such as Rank Loss, One Error, Hamming Loss, and Zero One Loss are also low for BCC/JRIP model for all samples of dataset, making it the best candidate for better performance given the label dependencies.

## 1. Introduction:

Machine Learning is a part of data mining which deals with automatically building models on large dataset from which interesting patterns can be learnt. Traditional single label classification is concerned with learning from a set of examples that are associated with single label $\lambda$ from a set of disjoint labels L, $|L|>1$. In multilabel classification, the examples are associated with a set of labels $Y \subseteq L$ (Trohidis, Tsoumakas, Kalliris, & Vlahavas, 2008). The problem of learning from multi label data has recently attracted significant attention, motivated from an increasing amount of new applications such as semantic annotation of images (Boutell, Luo, Shen, & Brown, 2004) (Zhang & Zhou, 2007) (Yang, Kim, & Ro, 2007) and video (Qi, et al., 2007) (Snoek, Worring, Van Germet, Geusebroek, & Smeulders, 2006), functional genomics (Clare & King, 2001) (Elisseeff & Weston, 2002) (Blockeel, Schietgat, Struyf, Dz?eroski, & Clare, 2006) (Cesa-Bianchi, Gentile, & Zaniboni, 2006) (Barutcuoglu, Schapire, & Troyanskaya, 2006), music categorizations into emotions (Li & Ogihara, Detecting emotion in music, 2003) (Li & Ogihara, Toward intelligent music information retrieval, 2006) (Wieczorkowska, Synak, & Ras, 2006) (Trohidis, Tsoumakas, Kalliris, & Vlahavas, 2008) and directed marking (Zhang, Burer, & Street, 2006).

Jafer et al., (Jafer, Matwin, & Sokolova, 2014) talk about various types of attributes related to personal privacy. The four types discussed are (i) Explicit Identifier, (ii) Quasi Identifier, (iii) Sensitive Identifier, (iv) Non–sensitive identifier. Explicit identifiers are the set of attributes from which the individuals can be explicitly identified. Explicit identifier includes SIN, Name, Insurance ID, which are directly linked to the individual. Quasi Identifiers (QI) refer to a set of attributes that when "combined", could be linked to external datasets and potentially breach the privacy. Race, gender, age are commonly present among QIs. Sensitive attributes includes salary, disease and so on, which corresponds to person-specific private information. Finally, remaining attributes that do not fall into any of the above categories are grouped as non-sensitive attributes.

A large growth in the availability and need of dataset containing personal information had increased the importance of protecting privacy of individuals. Datasets which originally contain personal health information can provide evident data to identify the person and their condition. To prevent and control identification of individuals, the dataset has to go through the process of de-identification, where the features relating to a person's personal and social part is removed without compromising on accuracy of the dataset. With respect to analyzing and learning demographics from data, to the best of our knowledge, there is no empirical work that uses multi-label classification to estimate the risk of patient identification.

Our experiments were conducted on newly released dataset collected from 130 US hospitals for years 1999-2008[1]. The dataset is introduced in (Strack, et al., 2014). The initial dataset consisted of 101,766 records. The dataset presents more than 50 features in medical setup representing patient demographic features such as race, gender, age, weight etc., and hospital features such as medical specialty, lab test results, diagnosis code. In the field of the diabetes research, most of the data mining work deals with finding if a person has diabetes or not; lesser amount of work has been done in reverse engineering, i.e. to identify the identity of persons who are likely to be diagnosed with diabetes. Straight forward problem of identifying patients with diabetes can be seen as a supervised binary classification, with a data record having one target label with 2 category values (Cortez & Morais, 2007) (Garcia, Lee, Woodard, & Titus, 1996) (Arrue, Ollero, & Matinez de Dios, 2000) (Dzeroski, Kobler, Gjorgioski, & Panov, 2006), whereas the reverse engineered problem of simultaneous learning of the demographic attributes (race, age, gender) belongs to multilabel classification. In multilabel classification, a data record has k, k >1, target labels. Each label can have different number of categorical or numerical values. For example, categorizing the patient to a list of infected allergies, as an individual may be allergic to more than one different allergy at a given time.

The reminder of the report is organized as follows. Section 2 presents multilabel classification problems. Section 3 presents performance evaluation metrics. Section 4 introduces the dataset. Section 5 presents the experimental setup and the empirical results. Conclusions and future work are drawn in Section 6.

# 2. Multi-Labeled Classification

Multilabel classification task has been studied in traditional database mining scenarios, where the problem is split into two steps where each label is learnt individually and then merged later. Most of the multilabel algorithms can be seen as an ensemble of binary label algorithms. Besides the concept of multilabel classification, multilabel learning introduces the concept of multilabel ranking (Brinker, Urnkranz, & Ullermeier, 2006). Multilabel ranking can be seen as a generalization of multiclass classification, where instead of predicting only a single label, it predicts the ranking of all labels.

[1] http://archive.ics.uci.edu/ml/datasets/Diabetes+130-US+hospitals+for+years+1999-2008#

Formally, let C be the set of instances and C ($c_1$, $c_2$, $c_3$…$c_n$, $x_i$), where $c_1$, $c_2$, $c_3$…$c_n$ represent the attributes and $x_i$ represents the target class label, for a given instance $c_i$ in C. In single label binary classification task $x_i = 2$, takes only two values, $c_i$ will belong to only one value of $x_i$. In single label multi class classification, $x_i > 2$, takes more than two values, $c_i$ belongs to one of the values of $x_i$. In multi label classification, there exists more than one target class $x_i$, $x_j$… $x_n$, such can each target label is binary and takes only values, each instance will belong to more than one target class. Multilabel classification methods can be categorized into two different groups (Trohidis & Tsoumakas, 2007): i) problem transformation methods, and ii) algorithm adaption methods. Problem transformation methods transform the problem of multilabel classification task into one or more single label classification, regression or ranking tasks. These group of methods are independent of the classification algorithm. Algorithm adaption category contains methods that extend specific single label learning algorithm to handle multiple data directly.

## 2.1 Problem transformation methods:

Problem transformation methods transform the problem of multilabel classification into a single label classification problem, they transform the data in such a way that existing single label algorithms can be applied. Transformation is done on the dataset so that the single label algorithms are applied. In problem transformation methods, the instance features are ignored, because they are not really important. In label power set transformation method, each unique set of labels in a multilabel training dataset is considered as one class in the new transformed data thus making the resulting dataset contain only one unique class for every entry. If the classifier can output a probability distribution over all new formed class, then it is possible to produce a ranking of labels (Read, 2008). Binary Relevance (BR) is one of the most popular approaches as a transformation that actually creates k datasets, k is the total number of classes, each for one class label and trains the classifier on each of these datasets. Each instance in the dataset $D_{\lambda j}$, $1 \leq j \leq k$, is either positively or negatively labelled, if they belong to class $\lambda_j$ and each of the datasets contain same number of instances as the original data. Binary classifier is trained for each of these datasets, once the datasets are transformed. BR assumes label independence for which it is implicitly criticized. Ranking by pairwise comparison transforms the multilabel dataset into binary label datasets, one for each pair of labels. Instance from the datasets, belonging to at least one of the corresponding labels but not both, are retained. Calibrated label ranking introduces the concept of introducing and additional label called calibration label, in the original dataset, to distinguish between the relevant and irrelevant labels. The calibration label can be seen as a neutral breaking point and all the labels that are less that the rank is treated as relevant set of labels, the ones that exceeds are considered as irrelevant. Each example that is annotated with a particular label, clearly is a positive example for that label and is treated as a negative example for the calibration label. Each example that is not annotated with a label is clearly a negative example for that label and is treated as a positive example for the calibration label.

## 2.2 Algorithm adaption methods:

Adaboost.MH and Adaboost.MR (Schapire, 2000) are two implementations based on tree boosting Adaboost algorithm, where Adaboost.MH tries to reduce hamming loss and the latter tries to find a hypothesis with optimal ranking. In Adaboost.MH, examples are presented as example label pairs and in each iteration increases the weights of misclassified example label pairs, but in contrast, AdaboostM.MR works on a pair of labels for any instance and in each iteration increases the weights of the example with mis-ordered label pairs. There exist many Lazy learning algorithms which are very similar, but the main differences occur in the way they aggregate the label sets for the given instances. BRkNN (Tsoumakas, T, Spyromitros, & Vlahavas, 2008) is a simple method which is logically equivalent to applying Binary Relevance followed by kNN. There are two main issued related to this method, computational complexity is multiplied by the number of labels and the other that none of the labels are included in at least half of the k nearest neighbors. BP-MLL is the common method that is based on Neural Network and Multilayer Perceptron based algorithms, where the error function (back propagation) has been modified to handle multilabel data. In this case, to handle multilabel data, one output is maintained for each class label. Multiclass Multilayer perceptron (MMP) proposed by Crammer and Singer in (Crammer & Kearns, 2003) leads to correct label ranking by updating the weight of the perceptron.

BR with SVM proposed by Godbole & Sarawagi (2004) proposed three ideas to improve the margin of overfitting for smaller training set. The first idea deals with having an extended dataset with K (=|L|) additional features so that BR will consider potential label dependencies, additional features are actually the predictions of each binary classifier at the first round. Confmat, is the second idea based on confusion matrix which removes negative training samples of a complete label if it is very similar to the positive label. The third idea is BandSVM, where on the learned decision hyperplane very similar negative examples are removed that are within a threshold distance and if there is a presence of overlapping classes, better models are built.

# 3 Performance Evaluation

In multi-labeled classification classifiers output a set of labels for every example, their prediction can be fully correct or partially correct or fully incorrect. Hence, single label accuracy metrics applied in their original form only partially capture multilabel performance and have to be adapted for multilabel problems (Sokolova, 2011). We present the evaluation measures used in our study.

Exact Match Ratio is an extension of accuracy of single label for multilabel prediction where partially correct and complete incorrect labels are treated as incorrect, since prediction of instances in multilabel data is a set of relevant and irrelevant labels and that the prediction can be fully correct or partially correct or fully incorrect.

$$ExactMatchRatio, MR = \frac{1}{n}\sum_{i=1}^{n} I(Y_i = Z_i)$$

I is the indicator function.

Precision is the proportion of predicted correct labels to the total number of actual labels, averaged across all the instances.

$$Precision, P = \frac{1}{n}\sum_{i=1}^{n}\frac{|Y_i \cap Z_i|}{|Z_i|}$$

Recall is the proportion of predicted correct labels to the total number of predicted labels, averaged over all instances.

$$Recall, R = \frac{1}{n}\sum_{i=1}^{n}\frac{|Y_i \cap Z_i|}{|Y_i|}$$

Accuracy is defined as the proportion of predicted correct labels to the total number of labels for that instance, averaged across all the instances.

$$Accuracy, A = \frac{1}{n}\sum_{i=1}^{n}\frac{|Y_i \cap Z_i|}{|Y_i \cup Z_i|}$$

$F_1$ score is defined as the harmonic mean of precision and recall, averaged over all the instances.

$$F_1 = \frac{1}{n}\sum_{i=1}^{n}\frac{2|Y_i \cap Z_i|}{|Y_i| + |Z_i|}$$

Hamming loss is the average of correctly predicted example to class ratio. If the HL is 0, then it would imply that there is no error, but it is nearly impossible, so smaller the value of HL, the better is the performance.

$$HammingLoss, HL = \frac{1}{kn}\sum_{i=1}^{n}\sum_{l=1}^{k}[I(l \in Z_i \wedge l \notin Y_i) + I(l \notin Z_i \wedge l \in Y_i)]$$

One-error is the count of how many times the top ranked predicted label is not in the set of true labels of the instance.

$$One - error, O = \frac{1}{n}\sum_{i=1}^{n} I(\arg\min_{\lambda \in \mathcal{L}} r_i(\lambda) \notin Y_i^l)$$

The top ranked predicted label is the label the classifier is most confident on and getting it wrong would clearly be an indication of overall lower performance of the classifier.

Ranking loss evaluates the average proportion of label pairs that are incorrectly ordered for an instance.

$$RankingLoss, RL = \frac{1}{n}\sum_{i=1}^{n}\frac{1}{|Y_i^l||\overline{Y_i^l}|}|(\lambda_a, \lambda_b) : r_i(\lambda_a) > r_i(\lambda_b), (\lambda_a, \lambda_b) \in Y_i^l \times \overline{Y_i^l}|$$

Similar to One-error, the smaller the ranking loss, the better the performance of the learning algorithm.

Coverage is the metric that evaluates how far on average a learning algorithm need to go down in order to cover all the true labels of an instance. Smaller the value of coverage, the better the performance.

$$Coverage, C = \frac{1}{n}\sum_{i=1}^{n}\max_{\lambda \in Y_i} r_i(\lambda) - 1$$

# 4 Data Set

In 2010, diabetes was the seventh leading cause of death mentioned in a total of 234,051 death certificates[2]. About 208,000 Americans under the age 20 are estimated to have diagnosed diabetes, which constitutes about 0.25% of the American population. In 2012, there were 1.7 million newly identified cases of diabetes. By population groups, 12.8%of Hispanics are diagnosed with diabetes. It is estimated that 7.6% and 13.2% is occupied by non-Hispanic black and non-Hispanic whites respectively. 9.0% of Asian Americans are affected by diabetes, while 15.9% of American Indians are diagnosed with diabetes. The dataset used in these experiments is extracted by Strack et. al. from national data warehouse that collects comprehensive clinical records across hospitals throughout the United States. The initial database consisted of data systematically collected from participating institutions and includes encounter data such as emergency, inpatient, and outpatient date. Patient demographics such as age, sex, and race were also present. Few medical parameters such as diagnosis code, in-hospital procedures documented by ICD-9-CM codes, laboratory results, in hospital mortality and hospital characteristics were also included. Dataset gathering was the combined work of Strack et. al. (2014) representing 10 years (1999-2008) of clinical care at 130 hospitals and integrated delivery networks throughout the United States.

[2] http://www.cdc.gov/diabetes/pubs/statsreport14/national-diabetes-report-web.pdf

Initial dataset consisted of 101767 records of male and female gender combined, each belonging to one of the following race: Caucasian, African American, Asian, Hispanic and other. The collection was created from large clinical database of diabetes patients based in the US hospitals. In the actual dataset, value of HbA1c was used as a marker of attention to identify if an individual is having a diagnosis of diabetes or not. 55 out of total 117 features were retained in the original dataset describing encounters, demographics, diagnoses, diabetic medications, number of visits and payment details. All the attribute information is briefly discussed in Table 1, Appendix A. In many researches of diabetic data mining and machine learning, the research entails on learning presence or absence of diabetes and HBA1c has great significance as diabetes marker and such learning process is a binary classification.

Our work involves learning demographic information of diabetes patients such as race and gender. Every male or female patient belonged to one of the following race groups: Caucasian, African American, Asian, Hispanic and Other. The age attribute which is grouped [(0-10), (10-20), etc.,] was not taken into current experiments as we considered that the process will be more complicated and beyond the scope of this project. On the pre-processing step, we removed several features that could not be treated directly since they had a high percentage of missing. These features were weight (97%), payer code (40%), and medical specialty (47%). Weight attribute was considered to be too sparse and it was not included in further analysis. Payer code was removed since it had a high percentage of missing values and it was not considered relevant to the outcome. Medical specialty attribute was maintained, adding the value "missing" in order to account for missing values. Personal Health Indicators (PHI) such as Encounter ID and Patient number were removed from the dataset as a step in de-identification process. Upon preprocessing and eliminating the features that were not related, we ended with 98054 instances and 45 attributes.

# 5 Empirical Results

Our experiments were conducted in three stages to find best classifier for learning demographics: the first stage consisted of applying several multi label classifiers on a sample dataset of first 1000 samples; the best performing algorithms were re-applied on second (10000 samples) and third (20000 samples) stages. We have chosen MEKA/ Mulan framework for running experiments. MEKA is based on WEKA framework. MEKA uses multiple attributes, one for each target label, where all variables are binary, indicating label relevance (1) or irrelevance (0), rather than a multi class – binary attribute. MULAN contains an evaluation framework that calculates a rich variety of performance measures; it is embedded into MEKA.

The first 1000 samples of the processed data were allocated for the initial set of experiments. Models were built on multilabel classifiers such as Binary Relevance (BR), Classifier Chains (CC), and Bayesian Classifier Chains (BCC), Binary Relevance quick (BRq), Conditional Dependency Network (CDN), and Majority Labeset. Based on accuracy and data type compatibility, we considered Binary Relevance, Classifier Chains and Bayesian Classifier Chains for further experiments. Using these three multi label classifiers as the base classifier (Step I) for transforming multilabel data into single label data, and for step 2, we decided on a few single label classifiers. (Recall that in MEKA each multilabel model is an ensemble of single label and multilabel classifier.) We ran Random Tree, Decision Tree, Decision Table, Multilayer Perceptron, Hoeffding Tree, k Nearest Neighbor, Naïve Bayes, JRIP. Testing time of Multilayer perceptron was longer than all other algorithms. The ones that were considered for further experiments are Random Tree, Decision Table, KNN, Hoeffding Tree, NaiveBayes, JRIP, and ZeroR. All the models were tested using test/ train split and Cross validation method and their results are tabulated in Table 1 below and individual values of accuracies and other measures are tabulated in Table 1 of Appendix A.

| **Model Name:** | **Test/Train Split** | **10 Fold CV** |
|---|---|---|
| | | |
| BR/Random Tree | 0.41 | 0.442 +/- 0.026 |
| CC/Random Tree | 0.489 | 0.438 +/- 0.019 |
| BCC/Random Tree | 0.44 | 0.464 +/- 0.024 |
| | | |
| BR/Decision Table | 0.51 | 0.513 +/- 0.031 |
| CC/Decision Table | 0.53 | 0.501 +/- 0.04 |
| BCC/Decision Table | 0.524 | 0.509 +/- 0.028 |
| | | |
| BR/KNN(5) | 0.52 | 0.523 +/- 0.025 |
| CC/KNN(5) | 0.513 | 0.51 +/- 0.034 |
| BCC/KNN(5) | 0.513 | 0.5 +/- 0.037 |
| | | |
| BR/Hoeffding Tree | 0.532 | 0.533 +/- 0.027 |
| CC/Hoeffding Tree | 0.512 | 0.52 +/- 0.03 |
| BCC/Hoeffding Tree | 0.512 | 0.521 +/- 0.03 |
| | | |
| BR/NaiveBayes | 0.488 | 0.525 +/- 0.04 |
| CC/NaiveBayes | 0.507 | 0.54 +/- 0.046 |
| BCC/NaiveBayes | 0.505 | 0.533 +/- 0.047 |
| | | |
| BR/JRIP | 0.479 | 0.509 +/- 0.031 |
| CC/JRIP | 0.528 | 0.532 +/- 0.034 |
| BCC/JRIP | 0.53 | 0.533 +/- 0.04 |

Table 1: Models and overall accuracies for 1000 samples,

BR/Hoeffding Tree and BCC/JRIP were chosen as the best algorithms based on their overall and individual accuracy measures. The results of these two models, from 10 fold cross validation method, is tabulated below in table 2.

| **Model Name:** | **Overall** | **1:Cauc** | **2:AfrAmer** | **3:Hisp** | **4:Asi** | **5:Oth** | **6:Male** | **7:Female** |
|---|---|---|---|---|---|---|---|---|
| BR/HoeffdingTree | 0.533 +/- 0.027 | 0.709 +/- 0.038 | 0.748 +/- 0.036 | 0.992 +/- 0.011 | 0.993 +/- 0.007 | 0.978 +/- 0.006 | 0.532 +/- 0.055 | 0.528 +/- 0.055 |
| BCC/JRIP | 0.533 +/- 0.04 | 0.712 +/- 0.054 | 0.74 +/- 0.056 | 0.991 +/- 0.01 | 0.993 +/- 0.007 | 0.978 +/- 0.006 | 0.524 +/- 0.048 | 0.524 +/- 0.048 |

Table 2: Models and accuracies for 1000 samples by using 10-fold CV

Once the results of the experiments were available for 1000 samples, further experiments were performed on two different randomly chosen bigger datasets of 10000 and 20000 samples respectively. Overall and individual accuracy for 10000 and 20000 samples are presented in Tables 3and 4.

| **Model** | **Overall** | **1:Cauc** | **2:AfrAmer** | **3:Hisp** | **4:Asi** | **5:Oth** | **6:Male** | **7:Female** |
|---|---|---|---|---|---|---|---|---|
| BR/ZeroR(baseline) | 0.586 +/- 0.009 | 0.792 +/- 0.012 | 0.83 +/- 0.011 | 0.982 +/- 0.004 | 0.994 +/- 0.003 | 0.986 +/- 0.004 | 0.543 +/- 0.013 | 0.543 +/- 0.013 |
| CC/ZeroR(baseline) | 0.586 +/- 0.009 | 0.792 +/- 0.012 | 0.83 +/- 0.011 | 0.982 +/- 0.004 | 0.994 +/- 0.003 | 0.986 +/- 0.004 | 0.543 +/- 0.013 | 0.543 +/- 0.013 |
| BCC/ZeroR(baseline) | 0.586 +/- 0.009 | 0.792 +/- 0.012 | 0.83 +/- 0.011 | 0.982 +/- 0.004 | 0.994 +/- 0.003 | 0.986 +/- 0.004 | 0.543 +/- 0.013 | 0.543 +/- 0.013 |
| | | | | | | | | |
| BR/HoeffdingTree | 0.624 +/- 0.014 | 0.784 +/- 0.013 | 0.789 +/- 0.028 | 0.97 +/- 0.005 | 0.985 +/- 0.006 | 0.976 +/- 0.003 | 0.629 +/- 0.019 | 0.637 +/- 0.022 |
| CC/HoeffdingTree | .628 +/- 0.013 | 0.772 +/- 0.014 | 0.813 +/- 0.016 | 0.978 +/- 0.004 | 0.994 +/- 0.003 | 0.985 +/- 0.003 | 0.631 +/- 0.023 | 0.631 +/- 0.023 |
| BCC/HoeffdingTree | 0.632 +/- 0.015 | 0.774 +/- 0.014 | 0.806 +/- 0.013 | 0.982 +/- 0.004 | 0.994 +/- 0.003 | 0.986 +/- 0.004 | 0.634 +/- 0.019 | 0.634 +/- 0.019 |
| | | | | | | | | |
| BR/JRIP | 0.693 +/- 0.023 | 0.796 +/- 0.015 | 0.836 +/- 0.009 | 0.982 +/- 0.005 | 0.994 +/- 0.003 | 0.986 +/- 0.004 | 0.703 +/- 0.088 | 0.728 +/- 0.01 |
| CC/JRIP | 0.704 +/- 0.012 | 0.798 +/- 0.011 | 0.835 +/- 0.01 | 0.982 +/- 0.004 | 0.994 +/- 0.003 | 0.986 +/- 0.004 | 0.728 +/- 0.017 | 0.728 +/- 0.017 |
| BCC/JRIP | 0.702 +/- 0.009 | 0.796 +/- 0.014 | 0.833 +/- 0.012 | 0.981 +/- 0.004 | 0.994 +/- 0.003 | 0.986 +/- 0.004 | 0.728 +/- 0.016 | 0.728 +/- 0.016 |

Table 3: Models and accuracies for 10000 samples using 10 fold CV

| **Model** | **Overall** | **1:Cauc** | **2:AfrAmer** | **3:Hisp** | **4:Asi** | **5:Oth** | **6:Male** | **7:Female** |
|---|---|---|---|---|---|---|---|---|
| BR/ZeroR(baseline) | 0.562 +/- 0.007 | 0.761 +/- 0.008 | 0.801 +/- 0.006 | 0.982 +/- 0.003 | 0.993 +/- 0.002 | 0.985 +/- 0.003 | 0.534 +/- 0.012 | 0.534 +/- 0.012 |
| CC/ZeroR(baseline) | 0.562 +/- 0.007 | 0.761 +/- 0.008 | 0.801 +/- 0.006 | 0.982 +/- 0.003 | 0.993 +/- 0.002 | 0.985 +/- 0.003 | 0.534 +/- 0.012 | 0.534 +/- 0.012 |
| BCC/ZeroR(baseline) | 0.562 +/- 0.007 | 0.761 +/- 0.008 | 0.801 +/- 0.006 | 0.982 +/- 0.003 | 0.993 +/- 0.002 | 0.985 +/- 0.003 | 0.534 +/- 0.012 | 0.534 +/- 0.012 |
| | | | | | | | | |
| BR/HoeffdingTree | 0.577 +/- 0.008 | 0.755 +/- 0.008 | 0.79 +/- 0.007 | 0.979 +/- 0.004 | 0.991 +/- 0.002 | 0.982 +/- 0.003 | 0.568 +/- 0.009 | 0.57 +/- 0.01 |
| CC/HoeffdingTree | 0.582 +/- 0.007 | 0.753 +/- 0.009 | 0.793 +/- 0.008 | 0.978 +/- 0.003 | 0.993 +/- 0.002 | 0.984 +/- 0.003 | 0.568 +/- 0.008 | 0.568 +/- 0.008 |
| BCC/HoeffdingTree | 0.582 +/- 0.007 | 0.752 +/- 0.007 | 0.792 +/- 0.003 | 0.981 +/- 0.003 | 0.993 +/- 0.002 | 0.984 +/- 0.003 | 0.57 +/- 0.008 | 0.57 +/- 0.008 |
| | | | | | | | | |
| BR/JRIP | 0.569 +/- 0.009 | 0.76 +/- 0.008 | 0.8 +/- 0.007 | 0.982 +/- 0.003 | 0.993 +/- 0.002 | 0.985 +/- 0.003 | 0.545 +/- 0.012 | 0.545 +/- 0.01 |
| CC/JRIP | 0.573 +/- 0.008 | 0.762 +/- 0.009 | 0.8 +/- 0.007 | 0.982 +/- 0.003 | 0.993 +/- 0.002 | 0.985 +/- 0.003 | 0.543 +/- 0.011 | 0.543 +/- 0.011 |
| BCC/JRIP | 0.569 +/- 0.012 | 0.76 +/- 0.008 | 0.798 +/- 0.007 | 0.982 +/- 0.003 | 0.993 +/- 0.002 | 0.985 +/- 0.003 | 0.535 +/- 0.013 | 0.535 +/- 0.013 |

Table 4: Models and accuracies for 20000 samples using 10 fold CV

Our empirical results showed that Hoeffding Trees (Geoff, Laurie, & Pedro, 2001) and JRip (Cohen, 1995) outperformed other algorithms. Hoeffding Tree is an incremental, anytime decision tree induction algorithm. HT exploit the fact that a small sample can often be enough to choose an optimal setting attribute. HT is a common decision tree variant that supports the idea by Hoeffding bound, which quantifies the number of observations. JRip implements a propositional rule learner. It is based on association rule with reduced error pruning (REP). In REP rules algorithms, the training data is split into a growing set and a pruning set. First, an initial rule set is formed the growing set, using some heuristic method. This growing set rule set are repeatedly pruned by applying one of a set of pruning operators. At each stage of simplification, the pruning operator chosen is the one that yields the greatest reduction of error on the pruning set. Table 5 below compares various performance measures of all multilabel models and datasets. Other than accuracy, other performance measures available from MEKA that were considered are Exact Match Ratio, Hamming Score, Harmonic Score, F1 Micro Average, Rank Loss, One Error, Hamming Loss and Zero One Loss. The latter four measures Rank Loss, One Error, Hamming Loss and Zero One Loss represent loss functions, i.e. lesser loss function indicates better algorithms. The relations are vice versa for Hamming Score, Harmonic Score, Exact Match Ratio, F1 Micro Average where the performance is better if the performance values are higher. Exact Match Ratio exemplifies the most difficult to achieve overall accuracy: it counts only examples with fully correctly predicted labels. Note that the best Exact Match Ratio is achieved by BCC / JRIP. BCC is based on Bayesian probability of Classifier Chains; it considers label dependencies, which can an advantage in studies of patient data. Loss functions such as Rank Loss, One Error, Hamming Loss, and Zero One Loss are also low for BCC/JRIP model for 1000 samples, making it the best candidate for better performance given the label dependencies.

| **Model** | **Exact Match** | **Hamming Score** | **Harmonic Score** | **F1 Micro Average** | **Rank Loss** | **One Error** | **Hamming Loss** | **Zero One Loss** |
|---|---|---|---|---|---|---|---|---|
| **BR / HT/ 10000** | 0.436 +/- 0.026 | 0.824 +/- 0.008 | 0.754 +/- 0.011 | 0.702 +/- 0.012 | 0.144 +/- 0.011 | 0.263 +/- 0.02 | 0.176 +/- 0.008 | 0.564 +/- 0.026 |
| **BR / JRIP / 10000** | 0.514 +/- 0.129 | 0.861 +/- 0.013 | 0.808 +/- 0.012 | 0.762 +/- 0.009 | 0.086 +/- 0.003 | 0.164 +/- 0.011 | 0.139 +/- 0.013 | 0.486 +/- 0.129 |
| **CC / HT/ 10000** | 0.485 +/- 0.017 | 0.829 +/- 0.007 | 0.748 +/- 0.011 | 0.7 +/- 0.012 | 0.191 +/- 0.008 | 0.23 +/- 0.014 | 0.171 +/- 0.007 | 0.515 +/- 0.017 |
| **CC / JRIP/ 10000** | 0.585 +/- 0.016 | 0.865 +/- 0.006 | 0.804 +/- 0.01 | 0.763 +/- 0.011 | 0.143 +/- 0.006 | 0.202 +/- 0.011 | 0.135 +/- 0.006 | 0.415 +/- 0.016 |
| **BCC / HT/ 10000** | 0.491 +/- 0.019 | 0.83 +/- 0.008 | 0.751 +/- 0.012 | 0.703 +/- 0.014 | 0.19 +/- 0.008 | 0.229 +/- 0.014 | 0.17 +/- 0.008 | 0.509 +/- 0.019 |
| **BCC/JRIP/ 10000** | **0.582 +/- 0.013** | **0.864 +/- 0.004** | **0.803 +/- 0.006** | **0.761 +/- 0.007** | **0.146 +/- 0.004** | **0.205 +/- 0.014** | **0.136 +/- 0.004** | **0.418 +/- 0.013** |
| | | | | | | | | |
| **BR / HT/ 20000** | 0.401 +/- 0.015 | 0.805 +/- 0.003 | 0.711 +/- 0.005 | 0.66 +/- 0.006 | 0.147 +/- 0.003 | 0.292 +/- 0.006 | 0.195 +/- 0.003 | 0.599 +/- 0.015 |
| **BR / JRIP / 20000** | 0.326 +/- 0.011 | 0.801 +/- 0.004 | 0.71 +/- 0.011 | 0.658 +/- 0.008 | 0.123 +/- 0.003 | 0.239 +/- 0.008 | 0.199 +/- 0.004 | 0.674 +/- 0.011 |
| **CC / HT/ 20000** | 0.428 +/- 0.01 | 0.805 +/- 0.003 | 0.713 +/- 0.005 | 0.659 +/- 0.006 | 0.217 +/- 0.005 | 0.25 +/- 0.009 | 0.195 +/- 0.003 | 0.572 +/- 0.01 |
| **CC / JRIP/ 20000** | 0.416 +/- 0.009 | 0.801 +/- 0.004 | 0.706 +/- 0.007 | 0.652 +/- 0.007 | 0.224 +/- 0.006 | 0.239 +/- 0.009 | 0.199 +/- 0.004 | 0.584 +/- 0.009 |
| **BCC / HT/ 20000** | 0.427 +/- 0.012 | **0.806 +/- 0.003** | **0.713 +/- 0.005** | **0.66 +/- 0.006** | 0.216 +/- 0.004 | 0.248 +/- 0.007 | **0.194 +/- 0.003** | **0.573 +/- 0.012** |
| **BCC/ JRIP/ 20000** | 0.411 +/- 0.016 | 0.798 +/- 0.006 | 0.702 +/- 0.009 | 0.647 +/-0.01 | 0.228 +/- 0.007 | 0.241 +/- 0.008 | 0.202 +/- 0.006 | 0.589 +/- 0.016 |

Table 5: Comparison of various performance measures of all multilabel models and datasets

On 10000 and 20000 samples, BR/Hoeffding Tree performed better than the majority class classification. All the CC and BCC models of Hoeffding Tree and JRIP considerably outperformed the baseline accuracy. This shows the overall good performance prediction accuracy of the models built using Hoeffding Tree and JRIP algorithms. Though the individual performance of all models on three datasets (1000, 10000, 20000 samples) is better than the baseline (ZeroR), one of the contrasting observation is that individual accuracy, overall accuracy, and values of most of the other evaluation parameters are boosted from small dataset of 1000 samples to 100000 samples, whereas inconsistent between 10000 to 20000 samples. This observation can be due to the inconsistency in the dataset, were large subset of samples belonged to one or two target labels. Since ZeroR only predicts the mode for nominal class labels, the underlying multilabel base classifier has no impact on the overall performance of the ZeroR multiclass ensemble, which is the reason why accuracy of all the ZeroR models for a given dataset it same. Overall accuracy, individual accuracy and values of all the performance measures are tabulated in Tables 3 and 4 of Appendix A respectively.

Although the performance of BCC / Hoeffding Tree model did not outperform BCC/ JRIP model**,** on a much larger dataset of 20000 samples, BCC / Hoeffding Tree model performed better overall in terms of Hamming Score, Harmonic Score, F1 Micro Average and loss functions such as Hamming loss and Zero One Loss values were lesser than all other models.

## 6 Conclusion and Future work

In this empirical study, we approached learning of patient model based on different race and gender groups as multi-label problem. We used the UCI Diabetics dataset to obtain empirical evidence. The data set consisted of over 100,000 records and more than 50 features in medical setup based on 130 US hospitals. The dataset included personal demographics, diagnoses code, lab results, etc. and hospital features such as medical specialty, lab test results, diagnosis code etc. Our target classification labels were diabetes patients' demographics such as race and gender. Using MEKA/MULAN, we applied multi-label learning algorithms BR/Hoeffding Tree, CC/Hoeffding Tree, BCC/Hoeffding Tree, BR/JRIP. CC / JRIP, BCC/ JRIP respectively. The results of the models built were evaluated and compared based on several multi-label performance measures. Experiments conducted on 1000, 10000, 20000 samples of the Diabetics data sets have shown that the BR/JRIP algorithm achieves Exact Match Ratio of 0.533 (1000 samples), 0.702 (10000 samples), 0.569 (20000 samples), improving over the baseline model ZeroR with accuracy of 0.526, 0.586, .562 respectively.

Our results can be further used in Privacy-Preserving Data Mining. The important factor to be considered when involved in removing personal information is to make sure that removed features have no or very less effect on the classification accuracy. Data anonymization is an important process before releasing the dataset to make sure that the individuals representing the dataset are anonymous and has been considered seriously in the past. LeFevre, DeWitt, & Ramakrishnan (2006) proposed a suite of greedy algorithms in order to address the K anonymization problem for a number of analysis tasks such as classification and regression analysis for single/multiple categorical and numerical target attribute(s) respectively. Byun, Bertino, & Li (2005) proposed a comprehensive approach for privacy preserving access control based on the notion of purpose. Xiong and Rangachari (2008) presented an application-oriented approach for data anonymization which considers the relative attribute importance for the target applications

As an extension to the work done we foresee experiments utilizing the entire dataset. The results of experiments on the entire dataset would give a better insight into how the patients are dispersed. It would also be an interesting task to predict the age demographic in addition to race and gender, as it would represent complete demographic of the patient. Since this is the first work of its kind, it would be great idea to build models using algorithms that is capable of handling label dependencies. Apart from those mentioned above, it would be ideal if the rate of classification of different target label groups can be computed. In addition, our application can be further integrated into Privacy-Preserving Data Mining, where they can be used to assess risk of identification of different patient groups.

# Appendix A:

| **Model Name:** | **Overall** | **1:Cauc** | **2:AfrAmer** | **3:Hisp** | **4:Asi** | **5:Oth** | **6:Male** | **7:Female** |
|---|---|---|---|---|---|---|---|---|
| | | | | | | | | |
| BR/Random Tree | 0.41 | 0.63 | 0.648 | 0.985 | 0.985 | 0.973 | 0.512 | 0.455 |
| BCC/Random Tree | 0.44 | 0.627 | 0.655 | 0.985 | 0.991 | 0.964 | 0.473 | 0.506 |
| CC/Random Tree | 0.489 | 0.582 | 0.648 | 0.988 | 0.979 | 0.967 | 0.576 | 0.576 |
| | | | | | | | | |
| BCC/Decision Table | 0.524 | 0.679 | 0.712 | 0.991 | 0.994 | 0.982 | 0.545 | 0.545 |
| BR/Decision Table | 0.51 | 0.679 | 0.724 | 0.991 | 0.994 | 0.982 | 0.536 | 0.545 |
| CC/Decision Table | 0.53 | 0.691 | 0.724 | 0.991 | 0.994 | 0.982 | 0.545 | 0.545 |
| | | | | | | | | |
| BCC/KNN(5) | 0.513 | 0.655 | 0.682 | 0.991 | 0.994 | 0.979 | 0.545 | 0.545 |
| BR/KNN(5) | 0.52 | 0.673 | 0.609 | 0.991 | 0.994 | 0.964 | 0.488 | 0.539 |
| CC/KNN(5) | 0.513 | 0.655 | 0.685 | 0.991 | 0.994 | 0.982 | 0.539 | 0.539 |
| | | | | | | | | |
| BR/Hoeffding Tree | 0.532 | 0.676 | 0.718 | 0.991 | 0.994 | 0.982 | 0.479 | 0.521 |
| BCC/HoeffdingTree | 0.512 | 0.682 | 0.715 | 0.991 | 0.994 | 0.982 | 0.521 | 0.521 |
| CC/Hoeffding Tree | 0.512 | 0.685 | 0.718 | 0.991 | 0.994 | 0.982 | 0.521 | 0.521 |
| | | | | | | | | |
| BCC/NaiveBayes | 0.505 | 0.652 | 0.679 | 0.985 | 0.979 | 0.945 | 0.527 | 0.527 |
| BR/NaiveBayes | 0.488 | 0.652 | 0.691 | 0.988 | 0.979 | 0.961 | 0.506 | 0.506 |
| CC/NaiveBayes | 0.507 | 0.667 | 0.694 | 0.985 | 0.979 | 0.967 | 0.509 | 0.509 |
| | | | | | | | | |
| CC/JRIP | 0.528 | 0.691 | 0.724 | 0.988 | 0.994 | 0.982 | 0.524 | 0.524 |
| BCC/JRIP | 0.53 | 0.697 | 0.73 | 0.991 | 0.994 | 0.979 | 0.533 | 0.533 |
| BR/JRIP | 0.479 | 0.682 | 0.724 | 0.991 | 0.994 | 0.982 | 0.491 | 0.482 |

Table 1: Overall and individual accuracies of target labels for first 1000 samples using test/train split CV method

| **Model Name:** | **Overall** | **1:Cauc** | **2:AfrAmer** | **3:Hisp** | **4:Asi** | **5:Oth** | **6:Male** | **7:Female** |
|---|---|---|---|---|---|---|---|---|
| | | | | | | | | |
| BR/Random Tree | 0.442 +/- 0.026 | 0.634 +/- 0.047 | 0.688 +/- 0.039 | 0.983 +/- 0.012 | 0.989 +/- 0.009 | 0.956 +/- 0.014 | 0.519 +/- 0.041 | 0.524 +/- 0.048 |
| BCC/Random Tree | 0.464 +/- 0.024 | 0.623 +/- 0.052 | 0.679 +/- 0.04 | 0.982 +/- 0.011 | 0.986 +/- 0.005 | 0.968 +/- 0.011 | 0.526 +/- 0.034 | 0.525 +/- 0.035 |
| CC/Random Tree | 0.438 +/- 0.019 | 0.608 +/- 0.048 | 0.696 +/- 0.036 | 0.986 +/- 0.016 | 0.989 +/- 0.006 | 0.958 +/- 0.012 | 0.519 +/- 0.028 | 0.495 +/- 0.04 |
| | | | | | | | | |
| BCC/Decision Table | 0.509 +/- 0.028 | 0.697 +/- 0.041 | 0.733 +/- 0.04 | 0.992 +/- 0.011 | 0.993 +/- 0.007 | 0.978 +/- 0.006 | 0.486 +/- 0.029 | 0.486 +/- 0.029 |
| BR/Decision Table | 0.513 +/- 0.031 | 0.698 +/- 0.041 | 0.74 +/- 0.032 | 0.992 +/- 0.011 | 0.993 +/- 0.007 | 0.978 +/- 0.006 | 0.534 +/- 0.048 | .503 +/- 0.039 |
| CC/Decision Table | 0.501 +/- 0.04 | 0.705 +/- 0.04 | 0.742 +/- 0.033 | 0.992 +/- 0.011 | 0.993 +/- 0.007 | 0.978 +/- 0.006 | 0.472 +/- 0.054 | 0.472 +/- 0.054 |
| | | | | | | | | |
| BCC/KNN(5) | 0.5 +/- 0.037 | 0.663 +/- 0.044 | 0.688 +/- 0.051 | 0.992 +/- 0.011 | 0.993 +/- 0.007 | 0.978 +/- 0.006 | 0.513 +/- 0.04 | 0.513 +/- 0.04 |
| BR/KNN(5) | 0.523 +/- 0.025 | 0.711 +/- 0.053 | 0.589 +/- 0.045 | 0.991 +/- 0.012 | 0.993 +/- 0.007 | 0.972 +/- 0.011 | 0.485 +/- 0.051 | 0.535 +/- 0.057 |
| CC/KNN(5) | 0.51 +/- 0.034 | 0.675 +/- 0.046 | 0.708 +/- 0.046 | 0.992 +/- 0.011 | 0.993 +/- 0.007 | 0.978 +/- 0.006 | .517 +/- 0.039 | 0.517 +/- 0.039 |
| | | | | | | | | |
| BR/Hoeffding Tree | 0.533 +/- 0.027 | 0.709 +/- 0.038 | 0.748 +/- 0.036 | 0.992 +/- 0.011 | 0.993 +/- 0.007 | 0.978 +/- 0.006 | 0.532 +/- 0.055 | 0.528 +/- 0.055 |
| BCC/HoeffdingTree | 0.521 +/- 0.03 | 0.709 +/- 0.038 | 0.746 +/- 0.036 | 0.992 +/- 0.011 | 0.993 +/- 0.007 | 0.978 +/- 0.006 | 0.51 +/- 0.056 | 0.51 +/- 0.056 |
| CC/Hoeffding Tree | 0.52 +/- 0.03 | 0.709 +/- 0.043 | 0.746 +/- 0.039 | 0.992 +/- 0.011 | 0.993 +/- 0.007 | 0.978 +/- 0.006 | .51 +/- 0.069 | 0.51 +/- 0.069 |
| | | | | | | | | |
| BCC/ NaiveBayes | 0.533 +/- 0.047 | 0.69 +/- 0.047 | 0.71 +/- 0.069 | 0.986 +/- 0.013 | 0.987 +/- 0.009 | 0.958 +/- 0.017 | 0.548 +/- 0.063 | 0.548 +/- 0.063 |
| BR/ NaiveBayes | 0.525 +/- 0.04 | 0.692 +/- 0.044 | 0.73 +/- 0.051 | 0.986 +/- 0.011 | 0.988 +/- 0.008 | 0.96 +/- 0.016 | 0.539 +/- 0.048 | 0.542 +/- 0.047 |
| CC/ NaiveBayes | 0.54 +/- 0.046 | 0.705 +/- 0.051 | 0.73 +/- 0.053 | 0.987 +/- 0.011 | 0.989 +/- 0.008 | 0.97 +/- 0.01 | 0.542 +/- 0.063 | 0.542 +/- 0.063 |
| | | | | | | | | |
| CC/JRIP | 0.532 +/- 0.034 | 0.712 +/- 0.048 | 0.746 +/- 0.045 | 0.991 +/- 0.01 | 0.993 +/- 0.007 | 0.977 +/- 0.008 | 0.521 +/- 0.054 | 0.521 +/- 0.054 |
| BCC/JRIP | 0.533 +/- 0.04 | 0.712 +/- 0.054 | 0.74 +/- 0.056 | 0.991 +/- 0.01 | 0.993 +/- 0.007 | 0.978 +/- 0.006 | 0.524 +/- 0.048 | 0.524 +/- 0.048 |
| BR/JRIP | 0.509 +/- 0.031 | 0.712 +/- 0.054 | 0.741 +/- 0.037 | 0.99 +/- 0.013 | 0.993 +/- 0.007 | 0.977 +/- 0.007 | 0.534 +/- 0.047 | 0.514 +/- 0.059 |

Table 2: Overall and individual accuracies of target labels for first 1000 samples using k-fold CV method

| Model | BR/ ZeroR | CC/ ZeroR | BCC/ ZeroR | BR/HT | CC/HT | BCC/ HT | BR/ JRIP | CC /JRIP | BCC/ JRIP | BR/ ZeroR | CC/ ZeroR | BCC/ ZeroR | BR/ HT | CC/ HT | BCC/ HT | BR/ JRIP | CC/ JRIP | BCC/ JRIP |
|---|---|---|---|---|---|---|---|---|---|---|---|---|---|---|---|---|---|---|
| | Test /Train Split CV method | | | | | | | | | 10 Fold CV method | | | | | | | | |
| **Overall** | 0.576 | 0.576 | 0.576 | 0.618 | 0.615 | 0.629 | 0.695 | 0.703 | 0.698 | 0.586 +/- 0.009 | 0.586 +/- 0.009 | 0.586 +/- 0.009 | 0.624 +/- 0.014 | .628 +/- 0.013 | 0.632 +/- 0.015 | 0.693 +/- 0.023 | 0.704 +/- 0.012 | 0.702 +/- 0.009 |
| **1:Cauc** | 0.788 | 0.788 | 0.788 | 0.783 | 0.75 | 0.773 | 0.79 | 0.79 | 0.79 | 0.792 +/- 0.012 | 0.792 +/- 0.012 | 0.792 +/- 0.012 | 0.784 +/- 0.013 | 0.772 +/- 0.014 | 0.774 +/- 0.014 | 0.796 +/- 0.015 | 0.798 +/- 0.011 | 0.796 +/- 0.014 |
| **2:AfrAmer** | 0.827 | 0.827 | 0.827 | 0.784 | 0.799 | 0.805 | 0.826 | 0.826 | 0.826 | 0.83 +/- 0.011 | 0.83 +/- 0.011 | 0.83 +/- 0.011 | 0.789 +/- 0.028 | 0.813 +/- 0.016 | 0.806 +/- 0.013 | 0.836 +/- 0.009 | 0.835 +/- 0.01 | 0.833 +/- 0.012 |
| **3:Hisp** | 0.981 | 0.981 | 0.981 | 0.958 | 0.949 | 0.981 | 0.981 | 0.981 | 0.981 | 0.982 +/- 0.004 | 0.982 +/- 0.004 | 0.982 +/- 0.004 | 0.97 +/- 0.005 | 0.978 +/- 0.004 | 0.982 +/- 0.004 | 0.982 +/- 0.005 | 0.982 +/- 0.004 | 0.981 +/- 0.004 |
| **4:Asi** | 0.992 | 0.992 | 0.992 | 0.967 | 0.992 | 0.992 | 0.992 | 0.992 | 0.992 | 0.994 +/- 0.003 | 0.994 +/- 0.003 | 0.994 +/- 0.003 | 0.985 +/- 0.006 | 0.994 +/- 0.003 | 0.994 +/- 0.003 | 0.994 +/- 0.003 | 0.994 +/- 0.003 | 0.994 +/- 0.003 |
| **5:Oth** | 0.988 | 0.988 | 0.988 | 0.965 | 0.988 | 0.988 | 0.988 | 0.987 | 0.988 | 0.986 +/- 0.004 | 0.986 +/- 0.004 | 0.986 +/- 0.004 | 0.976 +/- 0.003 | 0.985 +/- 0.003 | 0.986 +/- 0.004 | 0.986 +/- 0.004 | 0.986 +/- 0.004 | 0.986 +/- 0.004 |
| **6:Male** | 0.529 | 0.529 | 0.529 | 0.631 | 0.633 | 0.63 | 0.73 | 0.737 | 0.729 | 0.543 +/- 0.013 | 0.543 +/- 0.013 | 0.543 +/- 0.013 | 0.629 +/- 0.019 | 0.631 +/- 0.023 | 0.634 +/- 0.019 | 0.703 +/- 0.088 | 0.728 +/- 0.017 | 0.728 +/- 0.016 |
| **7:Female** | 0.529 | 0.529 | 0.529 | 0.635 | 0.633 | 0.63 | 0.726 | 0.737 | 0.729 | 0.543 +/- 0.013 | 0.543 +/- 0.013 | 0.543 +/- 0.013 | 0.637 +/- 0.022 | 0.631 +/- 0.023 | 0.634 +/- 0.019 | 0.728 +/- 0.01 | 0.728 +/- 0.017 | 0.728 +/- 0.016 |
| **Hamming Score** | 0.805 | 0.805 | 0.805 | 0.818 | 0.821 | 0.829 | 0.862 | 0.864 | 0.862 | 0.81 +/- 0.005 | 0.81 +/- 0.005 | 0.81 +/- 0.005 | 0.824 +/- 0.008 | 0.829 +/- 0.007 | 0.83 +/- 0.008 | 0.861 +/- 0.013 | 0.865 +/- 0.006 | 0.864 +/- 0.004 |
| **Exact Match** | 0.412 | 0.412 | 0.412 | 0.434 | 0.473 | 0.488 | 0.56 | 0.584 | 0.578 | 0.423 +/- 0.011 | 0.423 +/- 0.011 | 0.423 +/- 0.011 | 0.436 +/- 0.026 | 0.485 +/- 0.017 | 0.491 +/- 0.019 | 0.514 +/- 0.129 | 0.585 +/- 0.016 | 0.582 +/- 0.013 |
| **Hamming Loss** | 0.195 | 0.195 | 0.195 | 0.182 | 0.179 | 0.171 | 0.138 | 0.136 | 0.138 | 0.19 +/- 0.005 | 0.19 +/- 0.005 | 0.19 +/- 0.005 | 0.176 +/- 0.008 | 0.171 +/- 0.007 | 0.17 +/- 0.008 | 0.139 +/- 0.013 | 0.135 +/- 0.006 | 0.136 +/- 0.004 |
| **Zero One Loss** | 0.588 | 0.588 | 0.588 | 0.566 | 0.527 | 0.512 | 0.44 | 0.416 | 0.422 | 0.577 +/- 0.011 | 0.577 +/- 0.011 | 0.577 +/- 0.011 | 0.564 +/- 0.026 | 0.515 +/- 0.017 | 0.509 +/- 0.019 | 0.486 +/- 0.129 | 0.415 +/- 0.016 | 0.418 +/- 0.013 |
| **Harmonic Score** | 0.716 | 0.716 | 0.716 | 0.747 | 0.735 | 0.749 | 0.168 | 0.804 | 0.8 | 0.724 +/- 0.008 | 0.724 +/- 0.008 | 0.724 +/- 0.008 | 0.754 +/- 0.011 | 0.748 +/- 0.011 | 0.751 +/- 0.012 | 0.808 +/- 0.012 | 0.804 +/- 0.01 | 0.803 +/- 0.006 |
| **One Error** | 0.212 | 0.212 | 0.212 | 0.266 | 0.261 | 0.23 | 0.162 | 0.212 | 0.212 | 0.208 +/- 0.012 | 0.208 +/- 0.012 | 0.208 +/- 0.012 | 0.263 +/- 0.02 | 0.23 +/- 0.014 | 0.229 +/- 0.014 | 0.164 +/- 0.011 | 0.202 +/- 0.011 | 0.205 +/- 0.014 |
| **Rank Loss** | 0.117 | 0.208 | 0.117 | 0.143 | 0.193 | 0.188 | 0.086 | 0.143 | 0.145 | 0.115 +/- 0.004 | 0.115 +/- 0.004 | 0.115 +/- 0.004 | 0.144 +/- 0.011 | 0.191 +/- 0.008 | 0.19 +/- 0.008 | 0.086 +/- 0.003 | 0.143 +/- 0.006 | 0.146 +/- 0.004 |
| **F1 Micro Average** | 0.659 | 0.659 | 0.659 | 0.693 | 0.686 | 0.7 | 0.759 | 0.763 | 0.759 | 0.668 +/- 0.008 | 0.668 +/- 0.008 | 0.668 +/- 0.008 | 0.702 +/- 0.012 | 0.7 +/- 0.012 | 0.703 +/- 0.014 | 0.762 +/- 0.009 | 0.763 +/- 0.011 | 0.761 +/- 0.007 |

Table 3: Accuracies and other performance measures for 10000 samples (HT: Hoeffding Tree)

| Model | BR/ZeroR | CC/ZeroR | BCC/ZeroR | BR / HT | CC / HT | BCC / HT | BR/JRIP | CC / JRIP | BCC / JRIP | BR/ZeroR | CC / ZeroR | BCC/ ZeroR | BR / HT | CC / HT | BCC / HT | BR / JRIP | CC / JRIP | BCC / JRIP |
|---|---|---|---|---|---|---|---|---|---|---|---|---|---|---|---|---|---|---|
| | Test /Train Split CV method | | | | | | | | | 10 Fold CV Method | | | | | | | | |
| **Overall** | 0.565 | 0.565 | 0.565 | 0.577 | 0.579 | 0.581 | 0.555 | 0.557 | 0.556 | 0.562 +/- 0.007 | 0.562 +/- 0.007 | 0.562 +/- 0.007 | 0.577 +/- 0.008 | 0.582 +/- 0.007 | 0.582 +/- 0.007 | 0.56 +/- 0.009 | 0.573 +/- 0.008 | 0.569 +/- 0.012 |
| **1:Cauc** | 0.76 | 0.76 | 0.76 | 0.755 | 0.751 | 0.751 | 0.758 | 0.76 | 0.758 | 0.761 +/- 0.008 | 0.761 +/- 0.008 | 0.761 +/- 0.008 | 0.755 +/- 0.008 | 0.753 +/- 0.009 | 0.752 +/- 0.007 | 0.76 +/- 0.008 | 0.762 +/- 0.009 | 0.76 +/- 0.008 |
| **2:AfrAmer** | 0.801 | 0.801 | 0.801 | 0.79 | 0.793 | 0.789 | 0.8 | 0.8 | 0.797 | 0.801 +/- 0.006 | 0.801 +/- 0.006 | 0.801 +/- 0.006 | 0.79 +/- 0.007 | 0.793 +/- 0.008 | 0.792 +/- 0.003 | 0.8 +/- 0.007 | 0.8 +/- 0.007 | 0.798 +/- 0.007 |
| **3:Hisp** | 0.98 | 0.98 | 0.98 | 0.971 | 0.976 | 0.98 | 0.98 | 0.98 | 0.98 | 0.982 +/- 0.003 | 0.982 +/- 0.003 | 0.982 +/- 0.003 | 0.979 +/- 0.004 | 0.978 +/- 0.003 | 0.981 +/- 0.003 | 0.982 +/- 0.003 | 0.982 +/- 0.003 | 0.982 +/- 0.003 |
| **4:Asi** | 0.994 | 0.994 | 0.994 | 0.989 | 0.994 | 0.994 | 0.994 | 0.994 | 0.994 | 0.993 +/- 0.002 | 0.993 +/- 0.002 | 0.993 +/- 0.002 | 0.991 +/- 0.002 | 0.993 +/- 0.002 | 0.993 +/- 0.002 | 0.993 +/- 0.002 | 0.993 +/- 0.002 | 0.993 +/- 0.002 |
| **5:Oth** | 0.985 | 0.985 | 0.985 | 0.981 | 0.983 | 0.985 | 0.985 | 0.985 | 0.985 | 0.985 +/- 0.003 | 0.985 +/- 0.003 | 0.985 +/- 0.003 | 0.982 +/- 0.003 | 0.984 +/- 0.003 | 0.984 +/- 0.003 | 0.985 +/- 0.003 | 0.985 +/- 0.003 | 0.985 +/- 0.003 |
| **6:Male** | 0.539 | 0.539 | 0.539 | 0.564 | 0.564 | 0.566 | 0.545 | 0.514 | 0.515 | 0.534 +/- 0.012 | 0.534 +/- 0.012 | 0.534 +/- 0.012 | 0.568 +/- 0.009 | 0.568 +/- 0.008 | 0.57 +/- 0.008 | 0.545 +/- 0.012 | 0.543 +/- 0.011 | 0.535 +/- 0.013 |
| **7:Female** | 0.539 | 0.539 | 0.539 | 0.568 | 0.564 | 0.566 | 0.542 | 0.514 | 0.515 | 0.534 +/- 0.012 | 0.534 +/- 0.012 | 0.534 +/- 0.012 | 0.57 +/- 0.01 | 0.568 +/- 0.008 | 0.57 +/- 0.008 | 0.545 +/- 0.01 | 0.543 +/- 0.011 | 0.535 +/- 0.013 |
| **Hamming Score** | 0.8 | 0.8 | 0.8 | 0.803 | 0.804 | 0.804 | 0.801 | 0.792 | 0.792 | 0.798 +/- 0.004 | 0.798 +/- 0.004 | 0.798 +/- 0.004 | 0.805 +/- 0.003 | 0.805 +/- 0.003 | 0.806 +/- 0.003 | 0.801 +/- 0.004 | 0.801 +/- 0.004 | 0.798 +/- 0.006 |
| **Exact Match** | 0.395 | 0.395 | 0.395 | 0.402 | 0.423 | 0.428 | 0.325 | 0.399 | 0.397 | 0.392 +/- 0.009 | 0.392 +/- 0.009 | 0.392 +/- 0.009 | 0.401 +/- 0.015 | 0.428 +/- 0.01 | 0.427 +/- 0.012 | 0.326 +/- 0.011 | 0.416 +/- 0.009 | 0.411 +/- 0.016 |
| **Hamming Loss** | 0.2 | 0.2 | 0.2 | 0.197 | 0.196 | 0.196 | 0.199 | 0.208 | 0.208 | 0.202 +/- 0.004 | 0.202 +/- 0.004 | 0.202 +/- 0.004 | 0.195 +/- 0.003 | 0.195 +/- 0.003 | 0.194 +/- 0.003 | 0.199 +/- 0.004 | 0.199 +/- 0.004 | 0.202 +/- 0.006 |
| **Zero One Loss** | 0.605 | 0.605 | 0.605 | 0.598 | 0.577 | 0.572 | 0.675 | 0.601 | 0.603 | 0.608 +/- 0.009 | 0.608 +/- 0.009 | 0.608 +/- 0.009 | 0.599 +/- 0.015 | 0.572 +/- 0.01 | 0.573 +/- 0.012 | 0.674 +/- 0.011 | 0.584 +/- 0.009 | 0.589 +/- 0.016 |
| **Harmonic Score** | 0.708 | 0.708 | 0.708 | 0.712 | 0.71 | 0.711 | 0.702 | 0.692 | 0.691 | 0.706 +/- 0.006 | 0.706 +/- 0.006 | 0.706 +/- 0.006 | 0.711 +/- 0.005 | 0.713 +/- 0.005 | 0.713 +/- 0.005 | 0.71 +/- 0.011 | 0.706 +/- 0.007 | 0.702 +/- 0.009 |
| **One Error** | 0.24 | 0.24 | 0.24 | 0.315 | 0.251 | 0.251 | 0.242 | 0.241 | 0.243 | 0.239 +/- 0.008 | 0.239 +/- 0.008 | 0.239 +/- 0.008 | 0.292 +/- 0.006 | 0.25 +/- 0.009 | 0.248 +/- 0.007 | 0.239 +/- 0.008 | 0.239 +/- 0.009 | 0.241 +/- 0.008 |
| **Rank Loss** | 0.125 | 0.214 | 0.125 | 0.156 | 0.219 | 0.218 | 0.125 | 0.237 | 0.238 | 0.125 +/- 0.003 | 0.216 +/- 0.004 | 0.216 +/- 0.004 | 0.147 +/- 0.003 | 0.217 +/- 0.005 | 0.216 +/- 0.004 | 0.123 +/- 0.003 | 0.224 +/- 0.006 | 0.228 +/- 0.007 |
| **F1 Micro Average** | 0.649 | 0.649 | 0.649 | 0.658 | 0.656 | 0.658 | 0.653 | 0.637 | 0.636 | 0.647 +/- 0.006 | 0.647 +/- 0.006 | 0.647 +/- 0.006 | 0.66 +/- 0.006 | 0.659 +/- 0.006 | 0.66 +/- 0.006 | 0.658 +/- 0.008 | 0.652 +/- 0.007 | 0.647 +/- 0.01 |

Table 4: Accuracies and other performance measures for 20000 samples (HT: Hoeffding Tree)